\documentclass[10pt,twocolumn]{article}

\usepackage[T1]{fontenc}
\usepackage[utf8]{inputenc}
\usepackage[margin=1in]{geometry}
\usepackage[hyphens]{url}
\usepackage{graphicx}
\usepackage[numbers,sort&compress]{natbib}
\usepackage{caption}
\usepackage{algorithm}
\usepackage{algorithmic}
\usepackage{newfloat}
\usepackage{listings}
\usepackage{booktabs}
\usepackage{amsmath}
\usepackage{amssymb}
\usepackage{tabularx}
\usepackage[colorlinks=true,linkcolor=blue,citecolor=blue,urlcolor=blue]{hyperref}

\DeclareCaptionStyle{ruled}{labelfont=normalfont,labelsep=colon,strut=off}
\floatstyle{ruled}
\newfloat{listing}{tb}{lst}{}
\floatname{listing}{Listing}

\title{LabEvolver: Training-Free Experience Evolution for\\ Safe and Grounded Wet-Lab Agents}

\author{
Jingya Wang\textsuperscript{2}\thanks{Equal contribution.}
\quad
Yuyang Gao\textsuperscript{3}\footnotemark[1]
\quad
Liuzhenghao Lv\textsuperscript{3}
\\[0.25em]
Yonghong Tian\textsuperscript{1,2,3}\thanks{Corresponding author.}
\quad
Yuyang Liu\textsuperscript{1,2}\footnotemark[2]
\\[0.45em]
\small \textsuperscript{1}School of AI for Science, Peking University\\
\small \textsuperscript{2}School of Electronic and Computer Engineering, Peking University\\
\small \textsuperscript{3}School of Computer Science, Peking University\\[0.2em]
\small \texttt{\{liuyuyang13,yhtian\}@pku.edu.cn}
}

\date{}

\begin{document}

\maketitle

\begin{abstract}
We introduce \textbf{LabEvolver}, a training-free framework that equips safe and grounded wet-lab agents with episodic memory from execution experience. LabEvolver couples a state-grounded inner trial loop for adaptive perception, online planning, and safety validation with an outer evolution loop that distills completed trajectories into reusable skill, strategy, and safety experience. On robotic solution-preparation tasks, LabEvolver demonstrates real-world feasibility, reducing pH-regulation completion time and safety-gate intercepts by 48.2\% and 60.0\%, respectively. On ALFWorld, it further improves cumulative success rate within 20 steps from 76.2\% with ReAct to 91.4\% over 500 continual tasks, showing generality beyond wet-lab settings. These results support learn-by-doing experience evolution as a feasible path toward closed-loop automated scientific discovery. The project page is available at \url{https://andygao6186.github.io/LabEvolver/}.
\end{abstract}

\section{Introduction}
% Scientific discovery has evolved with the tools used to observe, explain, and explore the natural world. 
From empirical observation and theoretical modeling to computational simulation and data-driven science, successive paradigms have expanded the scale and speed of scientific discovery. The Fifth Paradigm, autonomous scientific discovery, is now becoming increasingly plausible~\cite{position}. Self-driving laboratories (SDLs) advance this vision by integrating robotics for high-throughput autonomous experiments~\cite{king2009automation,macleod2020self,burger2020mobile,szymanski2023autonomous}. However, most SDLs still depend on predefined interfaces and workflows, which lack a decision-making mechanism to understand high-level scientific goals and dynamically organize low-level operations~\cite{steiner2019organic,roch2020chemos}.

\begin{figure}[t]
    \centering
    \includegraphics[width=\columnwidth]{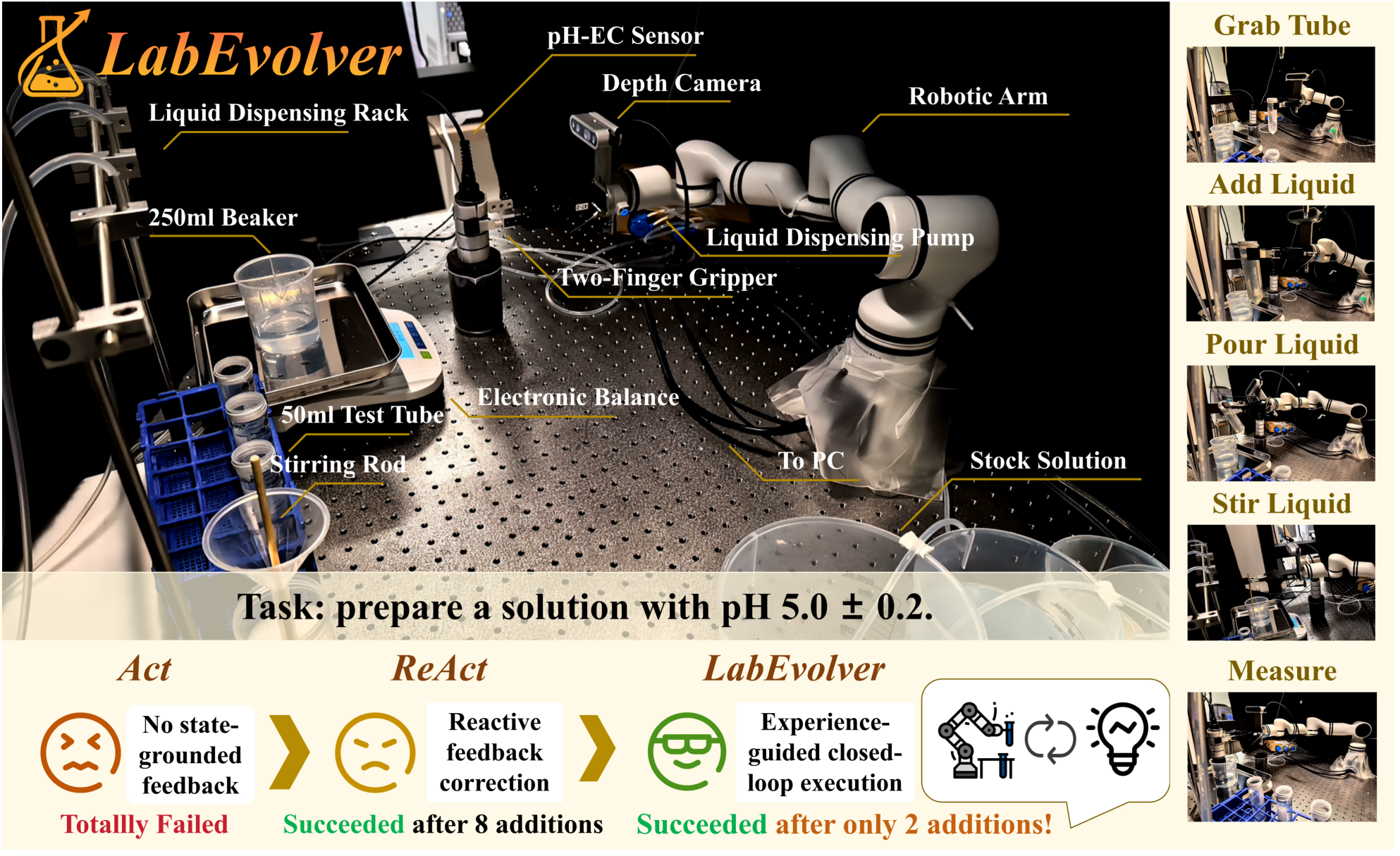} 
\caption{\textbf{Our proposed LabEvolver.}
In a pH-regulation task, LabEvolver surpasses one-shot action generation and purely within-trial feedback correction by leveraging accumulated experience.}
    \label{fig:hardware_platform}
\end{figure}

\begin{figure*}[t]
    \centering
    \includegraphics[width=\textwidth]{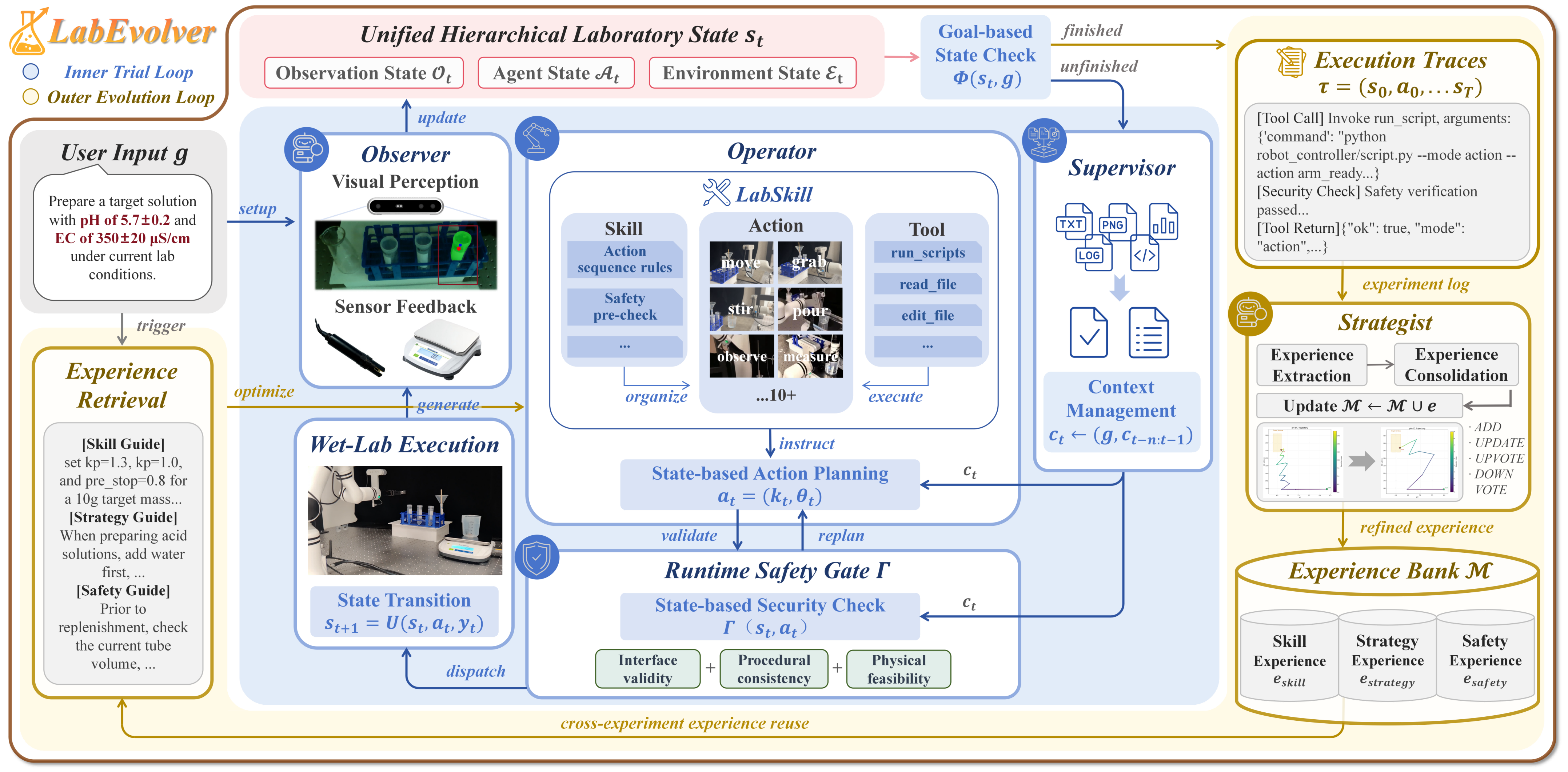} 
    \caption{\textbf{Overview of the LabEvolver Framework.} Given a high-level experimental goal, LabEvolver executes autonomous wet-lab workflows through a nested dual-loop process. (a) \textbf{\textit{Inner Trial Loop:}} the Observer maintains the hierarchical laboratory state $s_t$, the Operator maps goals to executable LabSkill actions under the tri-layer Safety Gate $\Gamma$, and the Supervisor manages runtime context for closed-loop replanning. (b) \textbf{\textit{Outer Evolution Loop:}} the Strategist distills completed execution traces into reusable skill, strategy, and safety experience, updating the global experience bank $\mathcal{M}$ for cross-experiment reuse.}
    \label{fig:overview}
\end{figure*}

In this context, foundation models offer a cognitive layer for SDLs by interpreting goals, generating plans, and connecting scientific knowledge with physical execution~\cite{boiko2023autonomous,m2024augmenting,zhao2025biomars}. However, many existing systems still represent physical actions as abstract symbols, leaving the planning and execution layers without a unified experimental state~\cite{boiko2023autonomous,m2024augmenting,yoshikawa2023large}. Vision-language-action (VLA) models further integrate perception, language, and robot actions, offering a promising route toward open-ended experimental operation~\cite{brohan2023rt2,kim2024openvla}. Yet their deployment in real scientific experiments remains bounded by practical limitations: 1) End-to-end policies often require costly scenario-specific data and are difficult to scale across diverse laboratory settings~\cite{brohan2023rt2,bharadhwaj2023roboagent,kim2024openvla}; 2) implicitly generated actions are difficult to trace, making their physical effects hard to predict or control under real-world constraints~\cite{darvish2025organa,ren2026labvla}.

These limitations highlight that embodied agents cannot rely solely on end-to-end action generation. Instead, they require grounded state anchors to plan physical actions prior to costly trial-and-error. This motivates a training-free, bounded framework that preserves flexible goal-directed control while enforcing explicit constraints in lab environments.

Beyond bounded execution within a single trial, autonomous laboratories must also improve across repeated experimental practice. Current embodied systems lack post-deployment practice loops, forcing each run to begin as a cold start without distilling trajectories into reusable knowledge~\cite{zhao2025biomars,angers2025roboculture,ren2026labvla}. Consequently, despite automating execution, they remain close to static intelligence. By contrast, human scientific expertise is not innate, but develops through conceptual understanding and hands-on practice~\cite{position}. To move toward the Fifth Paradigm of scientific discovery, autonomous laboratories should acquire a learn-by-doing capability that accumulates episodic memory from concrete operations~\cite{shinn2023reflexion,zhao2023expel,wang2023voyager}. This shifts the goal from static automation toward continuously learnable scientific discovery, where planning, cognition, and action improve together through practice.

To operationalize this paradigm, we introduce \textbf{LabEvolver}, an experience-driven and state-grounded autonomous experimentation framework that organizes laboratory workflows into a nested dual-loop process. In the inner trial loop, the system supports online execution through multi-agent collaboration under strict runtime safety gates. In the outer evolution loop, LabEvolver uses \textbf{Strategist} to distill completed experiments into reusable skills, strategies, and safety mitigation rules. Across real-world experiments, LabEvolver improves robust planning, safe execution, and cross-scenario evolution in long-horizon scientific tasks. Evaluation on ALFWorld further shows that LabEvolver provides a more general embodied cognitive architecture for experience reuse~\cite{ALFWorld20}.

Our contributions are summarized as follows:
\begin{itemize}
    \item We propose \textbf{LabEvolver}, a training-free dual-loop framework that integrates online closed-loop execution with post-trial experience evolution for autonomous scientific experimentation, without updating model weights.
    \item We introduce \textbf{Strategist}, a hierarchical experience manager that distills completed laboratory trajectories into reusable skill-level, strategy-level, and safety-level experience for better planning and execution.
    \item We validate LabEvolver through physical wet-lab experiments and ALFWorld, showing its effectiveness in safe execution, experience reuse, and cross-scenario transfer.
\end{itemize}

\section{Related Work}

\paragraph{Self-driving laboratories.}
Self-driving laboratories integrate robotic platforms, experimental design, and closed-loop optimization to accelerate scientific discovery. Prior work spans autonomous hypothesis testing~\cite{king2009automation}, programmable robotic workflows~\cite{steiner2019organic,roch2020chemos}, closed-loop discovery systems~\cite{macleod2020self,burger2020mobile,szymanski2023autonomous}, and recent wet-lab robotic platforms~\cite{zhao2025biomars,darvish2025organa,angers2025roboculture}. While existing systems establish the hardware foundation for automated experimentation, they typically require structured interfaces and rigid layouts. In contrast, LabEvolver addresses the complementary problem of flexible execution, enabling safe, state-grounded execution under changing wet-lab environments.

\paragraph{Foundation models for embodied experimentation.}
Foundation models increasingly serve as decision layers for robotic agents, supporting goal interpretation, tool use, and embodied planning~\cite{ahn2022saycan,liang2022codeaspolicies,singh2023progprompt,driess2023palme,boiko2023autonomous,m2024augmenting}. Robot foundation models and VLA policies connect language and vision to executable control~\cite{brohan2022rt1,openx2024rtx,khazatsky2024droid,brohan2023rt2,bharadhwaj2023roboagent,kim2024openvla,black2024pi0}, while laboratory-oriented agents extend these capabilities to scientific protocol execution~\cite{yoshikawa2023large,ren2026labvla}. However, many systems still depend on specialized interfaces or costly data. LabEvolver instead uses foundation models for high-level perception and planning, while grounding execution in structured laboratory states and LabSkill actions.

\paragraph{Self-evolving agents.}
Language and embodied agents can improve from past trials without updating model weights through reflection, procedural memory, and reusable skills~\cite{shinn2023reflexion,madaan2023selfrefine,zhao2023expel,fang2025memp,zhao2024learnact,zhang2025gmemory,xia2026skillrl,ni2026trace2skill}. In embodied settings, systems such as Voyager, EmbodiSkill, and ELITE accumulate experiential knowledge across tasks~\cite{wang2023voyager,ju2026embodiskill,wei2026elite}, with recent work extending self-evolution toward laboratory robotics~\cite{li2026roboclaw,huo2026abotclaw,wang2026autorobotist}. Most methods, however, remain in digital or general robotic settings where feedback is detached from deployed wet-lab constraints. LabEvolver addresses this gap by converting state-based wet-lab trajectories into practical skill, strategy, and safety experience.

\section{Methodology}

We formulate autonomous wet-lab experimentation as a constrained closed-loop problem. Given a high-level experimental goal $g$, LabEvolver seeks a physically viable trajectory $\tau$ that reaches the target experimental condition under real-world constraints. As shown in Figure~\ref{fig:overview}, LabEvolver adopts a nested dual-loop architecture. The inner trial loop constructs a hierarchical laboratory state, grounds goals into LabSkill actions, and validates each action through a tri-layer safety gate for state-grounded execution. The outer evolution loop leverages the Strategist to extract, maintain, and retrieve state-paired experience for future trials. Throughout the following exposition, we use the solution preparation task as a running example, where the goal $g$ is to bring the terminal solution pH to around 5.

\subsection{Inner Trial Loop: State-Driven Safe Execution}
The inner trial loop consists of three coupled mechanisms for hierarchical state construction, state-conditioned action planning, and runtime safety validation to support safe and robust long-horizon embodied wet-lab execution.

\subsubsection{Dynamic Hierarchical Laboratory State}
The shared laboratory state is the control anchor of the inner trial loop. LabEvolver maintains this state through the \textbf{Observer}, a perception module that converts free-form environmental feedback into structured variables for planning and safety checking. At step $t$, the system maintains
\[
s_t=(\mathcal{A}_t,\mathcal{E}_t,\mathcal{O}_t).
\]
Here, $\mathcal{A}_t$ denotes the agent embodiment state, including robot pose, gripper status, and interacting instruments. $\mathcal{E}_t$ denotes the environment state, including object identities, poses, and attributes. $\mathcal{O}_t$ denotes multimodal observations, including visual perception, sensor readings, and actuator feedback. 

\begin{figure}[tb]
\centering
\includegraphics[width=\columnwidth]{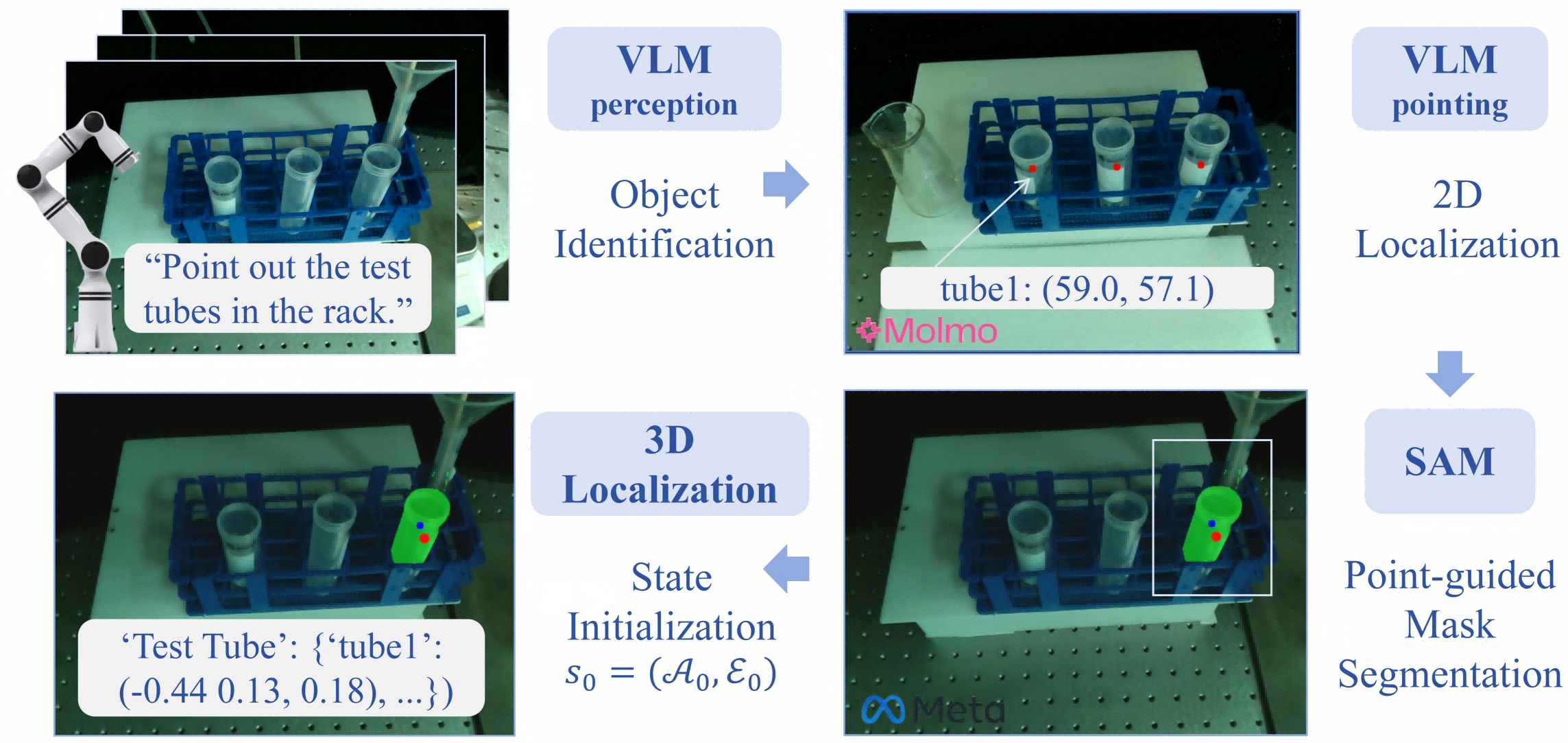}
\caption{\textbf{State Initialization Pipeline.} A four-stage process for the Observer to reconstruct the initial laboratory state $s_0$ from RGB-D data.}
\label{fig:state_initialization}
\end{figure}

Unlike prior robotic laboratory systems that rely on manually specified layouts~\cite{zhao2025biomars,darvish2025organa}, the Observer dynamically constructs this state from visual observations and runtime feedback. 

\begin{itemize}
    \item \textbf{State Initialization.} Before execution, the Observer reconstructs the initial state $s_0$ from RGB-D observations, as illustrated in Figure~\ref{fig:state_initialization}. It first identifies task-relevant objects through vision-language scene understanding, then localizes and segments these objects to estimate their spatial poses. The resulting identities, poses, and attributes are compiled into $\mathcal{E}_0=\{e_0^i=(\iota_0^i,p_0^i,x_0^i)\}_{i=1}^{N_0}$, while robot configuration initializes $\mathcal{A}_0=(q_0,h_0)$. For the running example, the perceived mixing beaker is represented as $e_0^{b}=(\iota_0^{b},p_0^{b},x_0^{b})$, with $x_0^{b}$ initialized as empty before preparation. Further robustness analysis of the Observer is provided in the supplementary document.
     \item \textbf{State Tracking.} During execution, the Observer updates $s_t$ from environmental feedback after each robotic action. These updates provide a reliable reference to the current laboratory state for subsequent planning, validation, and execution. During pH regulation, balance and pH-meter feedback update the beaker state $x_t^{b}$, including its solution mass and pH value.
\end{itemize}

\subsubsection{State-Conditioned Action Planning}
Based on the current laboratory state, the \textbf{Operator} uses LabSkill to ground high-level reasoning into executable laboratory actions while hiding hardware-specific control details. LabSkill is organized as a three-level action interface. The atomic action layer defines basic robot-executable primitives, the digital tool layer connects the Operator with external programs and runtime records, and the semantic skill layer provides task-level templates that compose atomic actions and tools into unified guidelines. This hierarchy allows the Operator to compose complex operations from reusable primitives, rather than following a hard-coded action sequence. See the supplementary document for more LabSkill details.

At step $t$, the Operator receives the planning context $c_t=(g,s_t,I_{\mathrm{LS}})$, where $g$ represents the experimental goal, $s_t$ the current laboratory state, and $I_{\mathrm{LS}}$ the LabSkill instructions. Conditioned on $c_t$, the Operator first generates a high-level intent $u_t$ and subsequently grounds it into a parameterized action $a_t$:
\[
u_t \sim \pi(c_t), \qquad a_t=(k_t,\theta_t)\sim \pi(\cdot \mid u_t,c_t),
\]
where $k_t$ denotes the action type, and $\theta_t$ specifies the target objects, waypoints, and interaction parameters dynamically determined by the Operator according to $s_t$ to match the laboratory environment. For example, when the latest pH measurement is above the pH 5 target range, the Operator may instantiate an acid-pouring action, with $\theta_t$ specifying the target beaker, pouring amount, and control parameters.

For further long-horizon execution, the \textbf{Supervisor} periodically compresses accumulated state transitions into structured progress checkpoints, reducing context saturation and goal drift.

\subsubsection{Tri-layer Runtime Safety Gate}
Because flexible LabSkill composition may produce actions that violate physical or procedural constraints, LabEvolver employs a \textbf{Tri-layer Runtime Safety Gate} as a safety boundary. Given the current state $s_t$ and a proposed action $a_t$, the gate validates the action at three levels before dispatch:
\[
\Gamma(s_t,a_t)=
\Gamma_{\mathrm{intf}}(s_t,a_t)
\wedge
\Gamma_{\mathrm{proc}}(s_t,a_t)
\wedge
\Gamma_{\mathrm{phys}}(s_t,a_t).
\]
Here, $\Gamma_{\mathrm{intf}}$ audits whether the LabSkill call is well formed, $\Gamma_{\mathrm{proc}}$ checks consistency with the current experimental progress, and $\Gamma_{\mathrm{phys}}$ verifies physical and measurement feasibility under $s_t$. For instance, $\Gamma_{\mathrm{proc}}$ blocks taking the pH meter when the robotic arm is still holding a test tube.

When $\Gamma(s_t,a_t)=1$, the action is dispatched to the physical platform, and the resulting feedback $y_{t+1}$ updates the state via $s_{t+1}=U(s_t,a_t,y_{t+1})$. Conversely, if $\Gamma(s_t,a_t)=0$, physical execution is halted immediately, and the gate generates a structured exception log identifying the violation to guide Operator replanning. By unifying pre-execution filtering with post-execution state updates, this mechanism establishes a reliable safety boundary for long-horizon wet-lab automation. A detailed analysis is provided in the supplementary document.

\subsection{Outer Evolution Loop: State-Paired Experience Refinement}
The outer evolution loop turns completed trials into state-paired experience fragments for future reuse. Instead of storing trajectories as stateless text logs, the Strategist links each reusable experience to the state condition under which it was useful, making retrieval sensitive to the current physical context rather than only to task keywords.

\paragraph{Experience extraction.}
After a trial terminates, the \textbf{Strategist} maps the trajectory $\tau$ and goal $g$ to a structured experience tuple $e=\Psi(\tau,g)=(e_{\mathrm{skill}},e_{\mathrm{strategy}},e_{\mathrm{safety}})$, where $e_{\mathrm{skill}}$ records effective LabSkill parameters and boundary conditions, $e_{\mathrm{strategy}}$ captures reusable procedural knowledge, and $e_{\mathrm{safety}}$ stores blocked-action diagnoses and failure-avoidance rules. In the running example, $e_{\mathrm{strategy}}$ can record that an initial pH of 7.74 suggests first adding about 8\,g of acid before re-measurement.

\paragraph{Experience maintenance.}
The extracted experience is merged into a global memory bank:
\[
\begin{aligned}
\mathcal{M} &\leftarrow \operatorname{Maintain}(\mathcal{M}, e, o), \\
o &\in \{\textsc{Add}, \textsc{Update}, \textsc{Upvote}, \textsc{Downvote}\}.
\end{aligned}
\]
By comparing newly extracted experience $e$ with existing records in $\mathcal{M}$, the Strategist inserts novel items, merges redundant entries, and adjusts memory priorities to retain useful records while suppressing unreliable ones. To keep long-term experience accumulation bounded, LabEvolver further applies an Ebbinghaus-style soft forgetting mechanism~\cite{ebbinghaus1913memory}, periodically decaying the scores of experience not reused in successful trials, and experience records with scores below zero are deleted. Details are provided in the supplementary document.

\paragraph{Experience retrieval.}
Before each trial, the Strategist retrieves top-$K$ relevant experience fragments from $\mathcal{M}$ based on the goal and initial state, forming a retrieved context $r$ to initialize the inner execution loop. During long-horizon tasks, retrieval is dynamically refreshed at Supervisor checkpoints. This state-based retrieval mechanism enables prior experience to guide both initial planning and mid-course replanning without replaying entire trajectories.

\begin{figure*}[t]
    \centering
    \includegraphics[width=\textwidth]{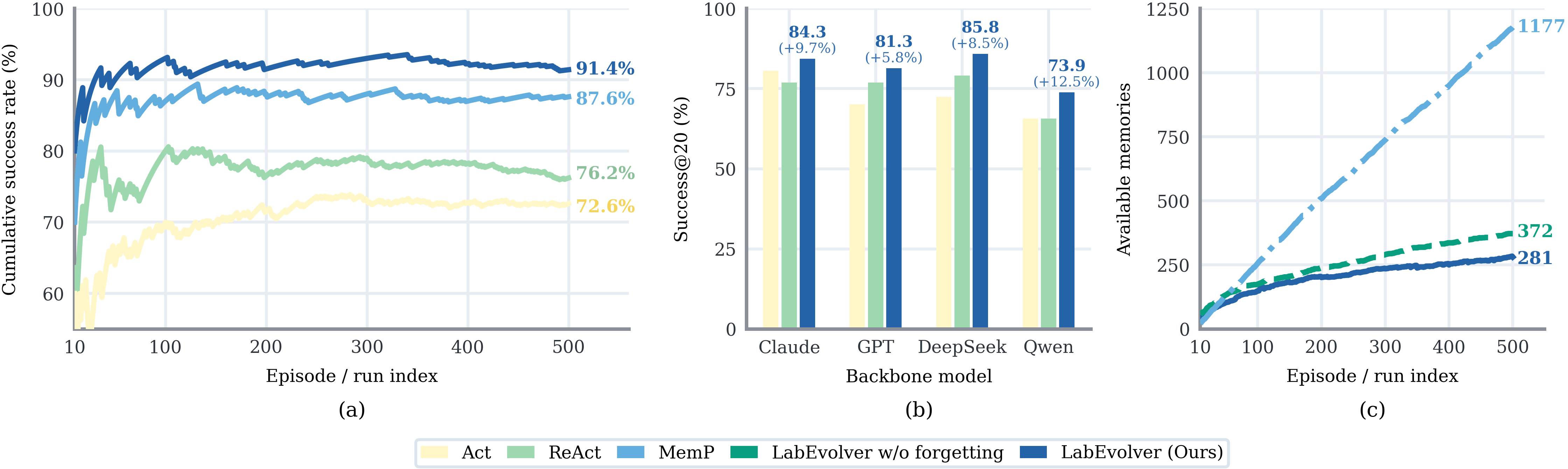}
\caption{ALFWorld results.
(a) Cumulative Success@20 over 500 continual tasks.
(b) Success@20 across backbone models on 134 \texttt{valid\_unseen} tasks.
(c) Available-memory growth over the 500-task continual stream.}
    \label{fig:alfworld_summary}
\end{figure*}

\section{Experiments}
\label{sec:experiments}

\subsection{Experimental Setup}
\label{sec:exp_setup}

We evaluate LabEvolver in two complementary regimes, progressing from simulation to real-world wet-lab validation. ALFWorld isolates outer-loop experience evolution in a scalable long-horizon benchmark to test algorithmic scalability and cross-task generalization. Robotic wet-lab tasks further assess the full dual-loop system under noisy physical feedback, state changes, and real-world safety constraints. Together, these evaluations test whether execution feedback becomes reusable experience while preserving state-valid laboratory operation.

\paragraph{Compared methods and backbones.}
Unless otherwise specified, we use DeepSeek-V4-Pro~\cite{deepseek2026deepseekv4} as the default backbone. Cross-model comparisons additionally include Claude-Sonnet-4.6~\cite{anthropic2026claude46}, Qwen3.5-35B-A3B~\cite{qwen2026qwen35b}, and GPT-5~\cite{openai2025gpt5}. We compare four methods: Act, which directly predicts actions; ReAct~\cite{yao2023react}, which adds within-task reasoning; Inner-only, which runs the complete inner loop without historical experience; and LabEvolver, which further implements outer-loop experience evolution.

\paragraph{ALFWorld benchmark.}
We evaluate scalable outer-loop experience evolution on long-horizon tasks using ALFWorld~\cite{ALFWorld20}. All experiments are training-free and conducted without any prior memory, treating the \texttt{train} and \texttt{valid\_unseen} splits purely as task streams. LabEvolver adopts the ReAct style prompt, incorporating retrieved experience as its sole additional context. We report the success rate within 20 steps (Success@20) across 134 \texttt{valid\_unseen} tasks, while tracking both task performance and experience accumulation metrics over a stream of 500 shuffled \texttt{train} tasks.

\paragraph{Robotic wet-lab platform.}
To validate the real-world effectiveness of LabEvolver, we conduct physical experiments on a wet-lab robotic platform shown in Figure~\ref{fig:hardware_platform}. The setup consists of a RealMan RM75-B arm, an Inspire Robots EG2-4B gripper, an Intel RealSense D435 RGB-D camera, an electronic balance, and a water-quality meter.

\paragraph{Wet-lab tasks and metrics.}
We evaluate three wet-lab task families: quantitative pouring, single-objective pH regulation, and coupled pH--EC regulation. As a core challenge in embodied manipulation, quantitative pouring tests whether physical feedback can be converted into reusable skill-level guidance. Serving as an indispensable foundation for scientific experimentation, solution preparation in pH and coupled pH--EC regulation evaluates experiment-level planning from sequential measurements, where the agent must select operations and action arguments online rather than follow a predefined protocol. With 11 action types, a $T$-step trajectory admits up to $11^T$ nominal action-type sequences, reaching $1.38\times10^{104}$ at $T=100$ before considering continuous parameters and state constraints. We report task success (Succ.), reagent additions (Add.), safety-gate intercepts (Gate), completion time, and task-specific accuracy.

\begin{figure*}[t]
    \centering
    \includegraphics[width=\textwidth]{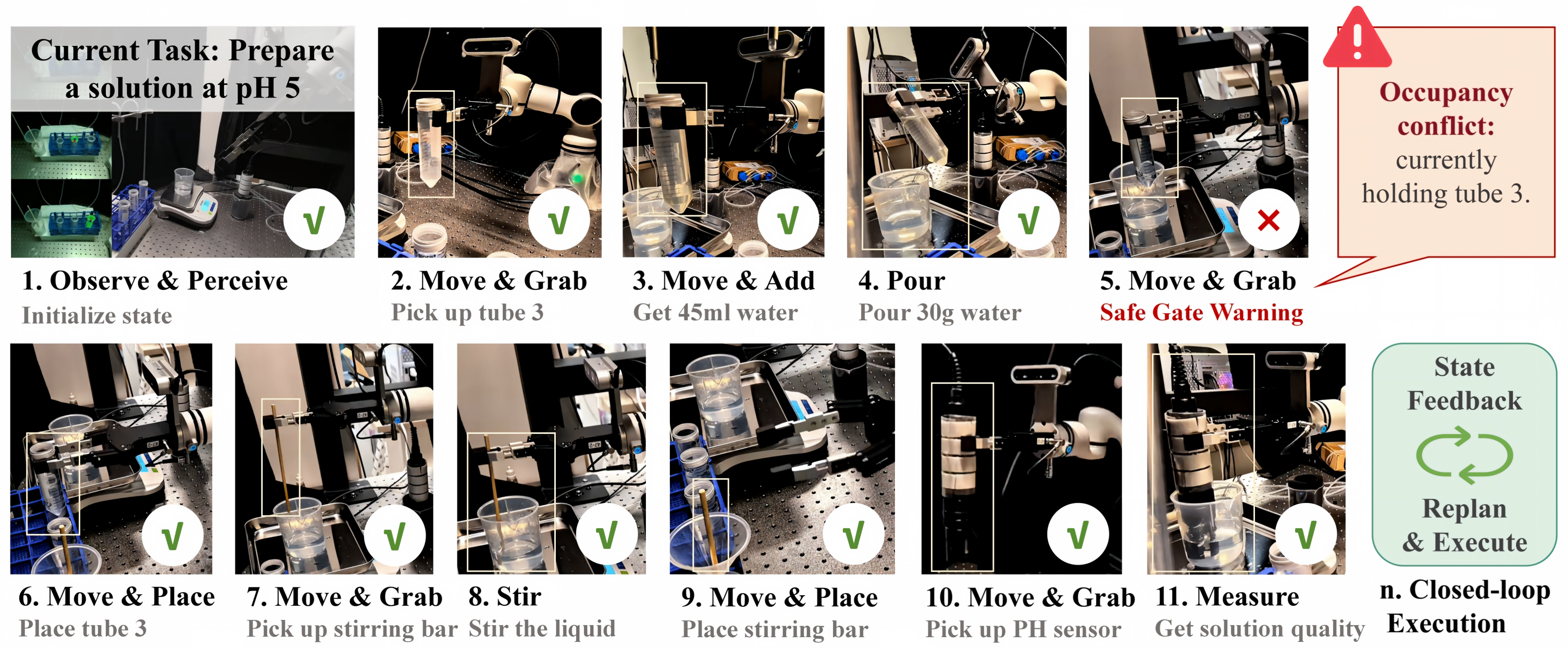}
    \caption{Representative pH regulation process showing safe closed-loop operation with state-feedback-driven replanning. Yellow boxes indicate the instruments currently manipulated by the robotic arm.}
    \label{fig:ph_closed_loop_execution}
\end{figure*}

\subsection{ALFWorld Results: Scalable Experience Improves Decision Making}
\label{sec:alfworld}

\paragraph{LabEvolver improves training-free long-horizon decision making.}
We first evaluate the outer evolution loop on 500 ALFWorld train-split episodes with DeepSeek-V4-Pro, comparing LabEvolver against Act, ReAct, and MemP~\cite{fang2025memp}, an online baseline that proceduralizes trajectories into workflows. As shown in Figure~\ref{fig:alfworld_summary}(a), LabEvolver achieves the highest cumulative success rate of 91.4\%, outperforming MemP by 3.8 points, ReAct by 15.2 points, and Act by 18.8 points. The large margins over non-memory baselines highlight the efficacy of test-time experience evolution, while the gain over MemP demonstrates that post-episode reflection and structured memory organization provide more actionable guidance than generic procedural workflows. Furthermore, LabEvolver's sustained advantage beyond early exploration indicates that its accumulated experience becomes increasingly reusable for downstream decisions.

\paragraph{LabEvolver generalizes across diverse backbone models.}
We next examine whether outer-loop experience improves decision making across different backbone models. On 134 \texttt{valid\_unseen} ALFWorld tasks, LabEvolver achieves the highest Success@20 for every tested backbone, as shown in Figure~\ref{fig:alfworld_summary}(b). The gain is smaller for stronger executors with already competitive action generation, such as Claude-Sonnet-4.6 and GPT-5, but becomes more pronounced for budget-sensitive models. For example, LabEvolver improves Qwen3.5-35B-A3B by 8.2 percentage points over ReAct and raises DeepSeek-V4-Pro from 72.4\% with Act to 85.8\%. These results suggest that retrieved experience reduces short-budget exploration and recovery costs, while the final gain depends on how effectively the executor interprets and applies the guidance. Full results, including cross-model and cross-method comparisons along with Success@50 metrics, are reported in the supplementary document.

\paragraph{LabEvolver prevents memory explosion.}
We further analyze memory-bank growth during continual ALFWorld evaluation. As shown in Figure~\ref{fig:alfworld_summary}(c), MemP grows almost linearly by proceduralizing every trajectory, reaching 1177 memories at episode 500. LabEvolver limits this growth to 372 memories through \textsc{Add}, \textsc{Update}, \textsc{Upvote}, and \textsc{Downvote}, which merge shared experience and admit only useful new fragments. With further long-term memory forgetting, the bank stays nearly flat after early accumulation and contains only 281 memories at episode 500. These results validate the memory-maintenance mechanism of the Strategist, which prevents memory explosion while compacting accumulated experience into a bounded memory bank. See the supplementary document for more detailed memory analysis.

\subsection{Pouring: Feedback-Driven Skill Evolution}
Quantitative pouring tests whether physical feedback can be converted into transferable LabSkill parameters. We implement pouring with a PD-based LabSkill whose $k_p$, $k_d$, and \texttt{pre\_stop} control wrist velocity and anticipatory stopping.

\paragraph{LabEvolver converts physical feedback into accurate pouring skills.}
We first evaluate the pouring LabSkill on a controlled 20\,g water-transfer task. We compare LabEvolver against three non-evolving controllers (fixed PD, manually tuned adaptive pre-stop PD, and CLAIRify~\cite{yoshikawa2023large}), repeating all comparisons three times to ensure consistency. Unlike these baselines, LabEvolver optimizes control parameters via outer-loop experience evolution across 40 self-exploration trials. As shown in Table~\ref{tab:pour20}, LabEvolver achieves the lowest MAE and shortest execution time. Compared with the manually tuned adaptive pre-stop PD controller, LabEvolver reduces MAE by 37.5\% and time by 45.1\%. This result shows that the outer loop can convert physical trials into a reusable control prior, while the inner loop dynamically grounds the evolved skill based on real-time mass measurements.

\begin{table}[t]
    \centering
    \small
    \setlength{\tabcolsep}{4pt}
    \begin{tabular}{lrrr}
        \toprule
        Method & MAE (g) $\downarrow$ & SD (g) $\downarrow$ & Time (s) $\downarrow$ \\
        \midrule
        Fixed PD & 0.500 & 0.100 & 21.035 \\
        Adaptive pre-stop PD & 0.133 & 0.153 & 20.360 \\
        CLAIRify & 0.867 & 0.379 & 61.930 \\
        LabEvolver (Ours) & \textbf{0.083} & 0.117 & \textbf{11.179} \\
        \bottomrule
    \end{tabular}
    \caption{Controller comparison for 20\,g water pouring.}
    \label{tab:pour20}
\end{table}

\begin{figure}[t]
    \centering
    \includegraphics[width=\columnwidth]{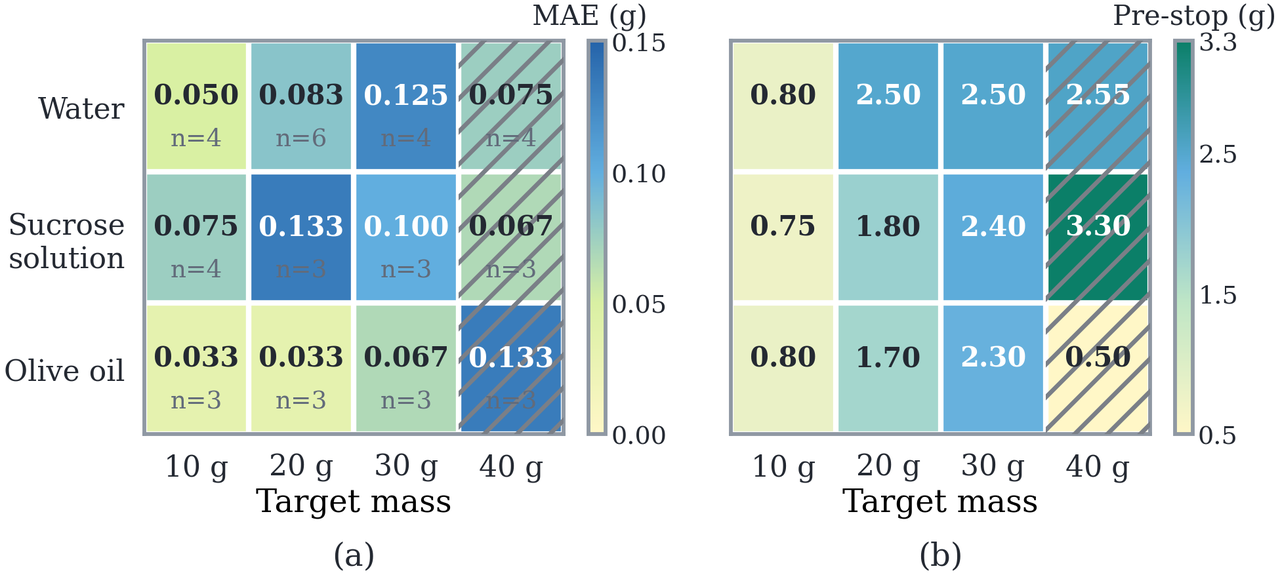}
\caption{Quantitative-pouring transfer across target masses and liquids.
(a) Mean absolute error and validation count for each locked-parameter condition.
(b) Adapted anticipatory stopping parameter across conditions.
Gray hatched regions denote conditions where the target mass approaches the test tube capacity.}
    \label{fig:pouring_summary}
\end{figure}

\paragraph{LabEvolver enables autonomous skill experience transfer.}
We next evaluate whether pouring experience transfers beyond the initial 20\,g water setting to water, sucrose solution, and olive oil at target masses of 10, 20, 30, and 40\,g. These conditions vary both target quantity and fluid dynamics, requiring adaptive pouring parameters. As shown in Figure~\ref{fig:pouring_summary}, 41 of 43 locked-parameter validation trials fall within $\pm0.2$\,g of the target mass, yielding a 95.3\% success rate and a trial-weighted MAE of 0.081\,g. LabEvolver automatically assigns condition-specific values of $k_p$, $k_d$, and \texttt{pre\_stop}, avoiding the manual retuning required by non-evolving controllers for each liquid-target pair. Full per-condition parameter details are provided in the supplementary document.

\subsection{pH Regulation: State-Grounded Strategy Evolution}
\label{sec:ph}

pH regulation tests whether state tracking and runtime safety gates keep long-horizon solution preparation valid, and whether strategy-level experience reduces unnecessary additions and invalid proposals.

\paragraph{LabEvolver achieves robust solution preparation across backbones.}
We compare the four agent configurations for pH 5 regulation across three backbone models. As shown in Table~\ref{tab:ph_backbones}, Act fails to reach the target across the tested backbones, indicating that one-shot action generation from the model's prior knowledge is insufficient for this task setting. ReAct and Inner-only recover task completion by using feedback in different ways, but both still incur extra additions or safety-gate interventions. By adding outer-loop experience, LabEvolver reduces the mean additions from 5.67 to 2.33 and completion time from 25.83 to 13.37 minutes relative to ReAct, while reducing mean gate intercepts from 6.00 to 0.67 relative to Inner-only under the same state mechanism. The improvement is larger for less reliable executors such as Qwen3.5-35B-A3B, indicating that retrieved experience can compensate for weaker planning. Overall, the inner loop supports robust online execution, while the outer loop improves efficiency and proposal quality.

\begin{table}[t]
    \centering
    \small
    \setlength{\tabcolsep}{3pt}
    \begin{tabular}{llcrrr}
        \toprule
        Backbone & Method & Succ. & Add. $\downarrow$ & Gate $\downarrow$ & Time (s) $\downarrow$ \\
        \midrule
        DeepSeek & Act & No & 4 & 0 & 1130 \\
        DeepSeek & ReAct & Yes & 8 & 1 & 1875 \\
        DeepSeek & Inner-only & Yes & 3 & 2 & 872 \\
        DeepSeek & LabEvolver & Yes & \textbf{2} & \textbf{0} & \textbf{652} \\
        \midrule
        Qwen & Act & No & 2 & 1 & 408 \\
        Qwen & ReAct & Yes & 4 & 4 & 706 \\
        Qwen & Inner-only & Yes & 7 & 15 & 2311 \\
        Qwen & LabEvolver & Yes & \textbf{2} & \textbf{2} & \textbf{609} \\
        \midrule
        Claude & Act & No & \textbf{2} & 1 & 476 \\
        Claude & ReAct & Yes & 5 & 0 & 2068 \\
        Claude & Inner-only & Yes & 4 & 1 & 1317 \\
        Claude & LabEvolver & Yes & 3 & \textbf{0} & \textbf{1145} \\
        \bottomrule
    \end{tabular}
    \caption{pH 5 regulation across DeepSeek-V4-Pro, Qwen3.5-35B-A3B, and Claude-Sonnet-4.6.}
    \label{tab:ph_backbones}
\end{table}

\paragraph{LabEvolver enables strategy experience transfer.}
We further examine repeated pH 5 regulation and transfer to pH 6 and pH 9 targets. LabEvolver consistently reaches the target ranges with limited operations, indicating that retrieved strategy experience can guide reagent selection and dose estimation while preserving measurement-conditioned replanning. Detailed results are provided in the supplementary document.

\subsection{Coupled pH--EC Regulation: Multi-Objective Evolution}
\label{sec:phec}

Unlike single-objective pH regulation, coupled pH--EC regulation requires the agent to coordinate two interacting solution properties, where different reagents can shift pH and EC in different directions with nonlinear and target-dependent effects. This task therefore evaluates how accumulated experience supports online planning under coupled multi-objective solution dynamics.

\begin{figure}[t]
    \centering
    \includegraphics[width=\columnwidth]{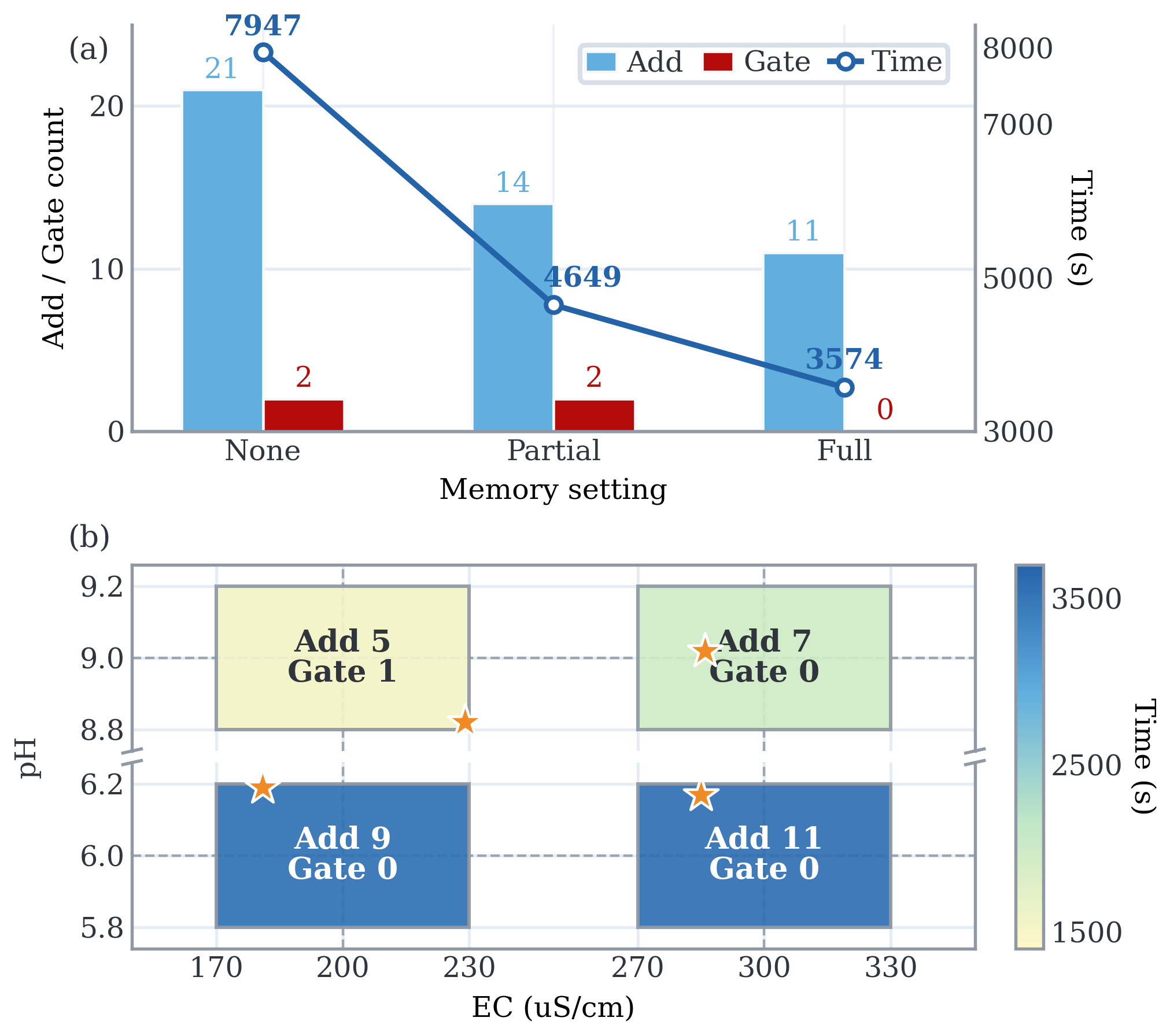}
    \caption{Coupled pH--EC regulation results.
    (a) Planning cost decreases as experience becomes more task-relevant.
    (b) Performance on coupled pH--EC targets. Rectangles denote admissible target regions, and stars denote final prepared states.}
    \label{fig:phec_summary}
\end{figure}

\paragraph{LabEvolver supports complex planning with task-relevant experience.}
We compare three experience settings on the same pH 6--EC 300 target to isolate the effect of experience relevance. The no-experience setting relies only on online feedback, the partial setting reuses experience from quantitative pouring and single-objective pH regulation, and the full setting further includes previous joint pH--EC trajectories while excluding the same target. As shown in Figure~\ref{fig:phec_summary}(a), partial experience already reduces exploratory operations, indicating that skill-level and single-objective strategy experience provide transferable priors for dosing and reagent selection. Full experience further reduces additions from 21 to 11, eliminates gate intercepts, and shortens completion time by 55.0\%, showing that joint pH--EC trajectories provide task-specific guidance for handling interacting objectives. Under the full experience setting, LabEvolver also reaches all four admissible pH--EC target regions in Figure~\ref{fig:phec_summary}(b), requiring 8 additions on average despite target-dependent and non-monotonic trajectories. These results suggest that beyond simply accumulating experience, LabEvolver further benefits from reusing history with higher task relevance, while relying on measurement-conditioned replanning during execution. Full per-target results are provided in the supplementary document.

\section{Conclusion}
We present \textbf{LabEvolver}, a training-free, dual-loop framework that enables flexible and safe robotic execution grounded in real-time laboratory states. By continually accumulating physical experience, it advances wet-lab agents from state-valid control to learn-by-doing autonomy, a practical step toward automated scientific discovery. Our wet-lab experiments demonstrate the real-world feasibility of this approach, while further evaluation in ALFWorld shows that its outer-loop experience distillation generalizes as a universal embodied cognition mechanism. Looking ahead, the long-horizon physical traces accumulated by LabEvolver offer valuable data to power future scientific world models. Despite these capabilities, current limitations mainly arise from relying on human-designed LabSkills and safety rules, which may constrain adaptation to unseen operations and novel failure modes.

\bibliographystyle{unsrtnat}
\bibliography{labevolver2027}

\end{document}